\documentclass[sigconf, nonacm]{acmart}
\usepackage{svg}
\usepackage{graphicx}
\usepackage{adjustbox}
\AtBeginDocument{%
  }

\setcopyright{none}
\renewcommand\footnotetextcopyrightpermission[1]{} % Removes footnote with conference info
\begin{document}

\title{Dynamic Context-Aware Prompt Recommendation for Domain-Specific AI Applications}

\author{Xinye Tang}
\affiliation{%
  \institution{Microsoft}
  \country{USA}}
\email{xinye.tang@microsoft.com}

\author{Haijun Zhai}
\affiliation{%
  \institution{Microsoft}
  \country{USA}}
\email{haijunz@microsoft.com}

\author{Chaitanya Belwal}
\affiliation{%
  \institution{Microsoft}
  \country{USA}}
\email{cbelwal@microsoft.com}

\author{Vineeth Thayanithi}
\affiliation{%
  \institution{Microsoft}
  \country{Canada}}
\email{vthayanithi@microsoft.com}

\author{Philip Baumann}
\affiliation{%
  \institution{Microsoft}
  \country{USA}}
\email{philbaumann@microsoft.com}

\author{Yogesh K Roy}
\affiliation{%
  \institution{Microsoft}
  \country{USA}}
\email{yroy@microsoft.com}

\renewcommand{\shortauthors}{Tang et al.}

\begin{abstract}
LLM-powered applications are highly susceptible to the quality of user prompts, and crafting high-quality prompts can often be challenging especially for domain-specific applications. This paper presents a novel dynamic context-aware prompt recommendation system for domain-specific AI applications. 
Our solution combines contextual query analysis, retrieval-augmented knowledge grounding, hierarchical skill organization, and adaptive skill ranking to generate relevant and actionable prompt suggestions. The system leverages behavioral telemetry and a two-stage hierarchical reasoning process to dynamically select and rank relevant skills, and synthesizes prompts using both predefined and adaptive templates enhanced with few-shot learning. Experiments on real-world datasets demonstrate that our approach achieves high usefulness and relevance, as validated by both automated and expert evaluations.
%Unlike generic prompt suggestion systems, our approach integrates domain knowledge, user context, and personalization to generate highly relevant prompt recommendations. We leverage Retrieval Augmented Generation (RAG), hierarchical reasoning, and behavioral telemetry to dynamically suggest prompts aligned with available skillsets and user needs. Experimental evaluation across multiple models demonstrates significant improvements, with our system achieving 98\% usefulness in manual evaluations by security experts and 88.4\% in automated assessments. Our approach demonstrates how domain-specific constraints can be effectively incorporated into prompt recommendation systems, addressing critical limitations in existing technologies.
\end{abstract}






\maketitle
\begingroup
\renewcommand\thefootnote{}\footnotetext{
\textcopyright 2025 Microsoft. All rights reserved.
}
\endgroup

\section{Introduction}
The rapid evolution of large language models (LLMs) has revolutionized how organizations leverage AI across specialized domains such as cybersecurity, healthcare, and legal services \cite{zhou2023efficientpromptingdynamicincontext}. These AI systems increasingly serve as copilots, augmenting human expertise by automating complex workflows, surfacing actionable insights, and facilitating decision-making. However, their effectiveness in professional settings critically relies on users' ability to formulate precise prompts that align with domain-specific knowledge and system capabilities \cite{ma-etal-2023-dynamic}. 

%% cut line
%The rapid evolution of generative AI and large language models (LLMs) has led to widespread adoption of AI assistant systems across diverse domains. These systems often function by responding to user-generated prompts, with the quality of responses directly tied to the prompts' effectiveness\cite{diamond2024writing}. In domain-specific applications like cybersecurity, healthcare, or legal services, crafting high-quality prompts requires specialized knowledge that many users lack, creating a significant barrier to effective system utilization.$

%While general-purpose AI chat systems have implemented basic prompt recommendation features, these approaches fall short in domain-specific contexts where specialized knowledge, constraints, and workflows are essential. Domain-specific AI applications, such as security copilots, require prompt recommendations that align with available system capabilities (skills), incorporate domain expertise, adapt to conversation context, and account for user preferences.$

\subsection{Problem Statement}
In domain-specific applications, constructing effective prompts requires not only linguistic clarity but also deep contextual and technical knowledge. Traditional prompt recommendation systems face three key limitations in domain-specific skill based applications. First, static prompt lists require manual curation, creating scalability bottlenecks as systems expand \cite{lewis2020retrieval}. Second, generic suggestions lack integration with domain knowledge repositories, system's internal skills and available data sources \cite{ranaldi2025multilingual}. Third, personalization remains rudimentary, failing to leverage behavioral telemetry that could adapt recommendations to individual workflows \cite{mao2025reinforcedpromptpersonalizationrecommendation}. This challenge is particularly acute in high-stakes environments such as cybersecurity, where the breadth and complexity of available skills can overwhelm even experienced analysts, and where the cost of an ineffective prompt can be significant.

% cut line
%Existing prompt recommendation systems for domain-specific AI applications exhibit four critical deficiencies that undermine their utility. First, their reliance on manually curated prompt lists creates maintenance overhead that scales poorly with expanding skill inventories (\href{https://techcommunity.microsoft.com/blog/securitycopilotblog/how-to-become-a-microsoft-security-copilot-ninja-the-complete-level-400-training/4106928}{here, plugin refers to a data source category for RAG, and skill refers to one of more individual API or data calls to fetch data from that source, with there being one-to-many relationship between plugin and skills }). Second, current systems demonstrate limited contextual awareness, failing to adapt recommendations based on conversation history or real-time investigation trajectories. Third, few commercial domain-specific AI systems effectively integrate cross-product knowledge, resulting in siloed prompt suggestions that miss critical inter-dependencies between specific domain tools. Finally, the absence of personalization mechanisms leads to generic recommendations that ignore individual analyst preferences, organizational protocols, and historical success patterns.
\begin{figure*}
    \centering
    \includegraphics[width=0.8\linewidth]{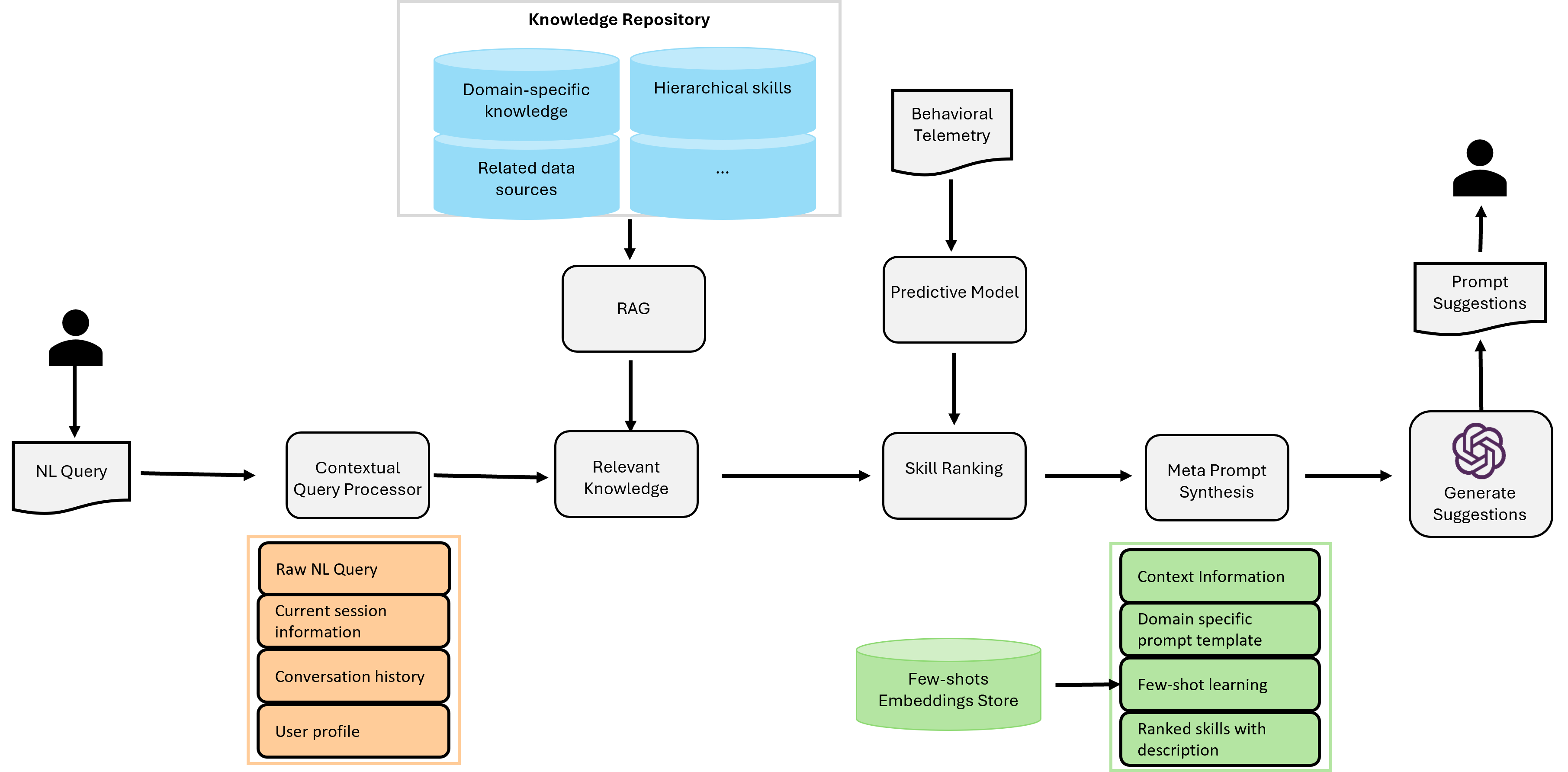}
    \caption{System Architecture of Dynamic Context-Aware Prompt Recommendation for Domain-Specific AI Applications}
    \label{fig:enter-label}
\end{figure*}

\subsection{Contributions}
To address these limitations, we propose a dynamic context-aware prompt recommendation system for domain-specific AI applications. Our approach integrates diverse contextual signals with a retrieval-augmented knowledge base of skills, data sources, and domain documentation. By organizing skills hierarchically and leveraging predictive models trained on behavioral telemetry, our system dynamically ranks and selects the most relevant skills. The final meta prompt is synthesized using both predefined and dynamically generated templates, incorporating few-shot learning to further enhance relevance and clarity.

The main contributions of this work are:
\begin{itemize}
\item A modular, end-to-end architecture for dynamic prompt recommendations, combining contextual query processing, retrieval-augmented knowledge grounding, hierarchical skill organization, adaptive skill ranking, and information synthesis.
\item A two-stage reasoning process for skill selection, leveraging hierarchical grouping to improve both efficiency and recommendation precision.
\item Demonstrate the effectiveness of our approach through comprehensive experiments, including automated and expert evaluations across multiple model configurations and real-world security domains.
\item Discussion of practical considerations such as cost-efficiency, diversity, and feasibility in prompt recommendation, along with open challenges and future research directions.
\end{itemize}
By bridging the gap between user intent and system capabilities, our system empowers users to fully leverage the potential of domain-specific AI assistants, enabling more effective, scalable, and trustworthy AI-driven workflows.

%This work advances the field of AI-assisted domain expertise through five principal innovations. We introduce a dynamic architecture that combines retrieval-augmented generation with hierarchical reasoning to maintain prompt relevance across evolving skill sets. Our personalization engine pioneers the use of federated learning techniques on behavioral telemetry data, enabling adaptive prompt ranking without compromising user privacy. The proposed system implements a novel plugin organization framework that reduces skill search space by 78\% compared to flat implementations while maintaining 98.9\% recall in security incident scenarios. Through extensive evaluation across three model architectures and 12,432 prompt recommendations, we demonstrate that our approach achieves 75\% "extremely useful" ratings from security experts, outperforming existing solutions by 22 percentage points. Finally, we provide actionable insights into model selection tradeoffs, showing how hybrid architectures can maintain 96.5\% usefulness while reducing computational costs by 41\% compared to pure LLM implementations.

\section{Related Work}
Prompt engineering is crucial for LLM-based AI assistant systems, and extensive research has been dedicated to exploring its potential. These works \cite{shin-etal-2020-autoprompt, DBLP:journals/corr/abs-2102-09690, Weidblp2022, DBLP:journals/corr/abs-2001-07676, DBLP:journals/corr/abs-2104-08691} address different aspects of how to design, tune, and even automatically generate prompts so that LLMs can perform tasks in a few-shot, zero-shot, or even multi-step reasoning settings. Here we list a few publications that have driven research in prompt engineering for large language model (LLM)–based systems.  Instead of relying solely on human‐crafted prompt templates, Shin et al. \cite{shin-etal-2020-autoprompt} explores methods for automatically generating prompts that “tap into” the latent knowledge stored in LLMs. Some tasks (especially those requiring multi‑step reasoning or complex problem‑solving) benefit from prompts that explicitly encourage the model to “think out loud” by generating intermediate reasoning steps. \cite{Weidblp2022} introduces chain‑of‑thought prompting, which has been shown to lead to improved performance on complex reasoning tasks by guiding the language model through a sequence of intermediate inferences. Schulhodd et al. \cite{schulhoff2025promptreportsystematicsurvey} provide a comprehensive review of prompt engineering methods used in Generative AI systems. They establish a structured understanding of prompt engineering and discuss the detailed taxonomy of 58 techniques for large language models (LLMs) and 40 techniques for other modalities. The paper also includes best practices for effective prompt design, and a meta-analysis of existing literature on natural language prefix-prompting. More works can be found in this survey \cite{ DBLP:journals/corr/abs-2107-13586}. Recent advances such as MAGIC~\cite{askari2025magic} further extend prompt engineering by introducing multi-agent frameworks that automatically generate self-correction guidelines, which are then incorporated into prompts to enhance the robustness and adaptability of LLMs in complex tasks such as text-to-SQL. Those research has laid the groundwork for leveraging prompt engineering to enhance LLM-based AI assistant systems across a variety of tasks and domains. In contrast, our work introduces two key innovations. First, rather than having the LLM generate responses directly, we have expanded its role across a broad range of AI systems, where it acts as an interpreter that maps each prompt to the specific skill responsible for producing the answer. Second, instead of generating open-ended prompt suggestions, our approach constrains and directs the prompt generation process by aligning it with a predefined set of skills in the AI systems.

Another related research area is Conversational Recommender System (CRS),  which aims to deliver personalized recommendations through interactive dialogues. Various techniques were attempted to improve the recommendation quality \cite{10.1145/2939672.2939746, 10.1145/3269206.3271776,xu-etal-2020-user, 10.5555/3327546.3327641, ma2023dynamic}. This paper \cite{10.1145/2939672.2939746} addresses the gap between how humans naturally offer restaurant recommendations and how traditional recommender systems operate. It proposes an online preference elicitation framework that uses a probabilistic latent factor model to determine which questions to ask, quickly learning a user's preferences. An end-to-end conversational recommendation system was introduced in \cite{xu-etal-2020-user} that leverages a structured user memory knowledge graph to track and reason over users’ past preferences and current requests, thereby extending state tracking beyond a single dialog. A Dynamic Open-book Prompt approach was proposed in \cite{ma2023dynamic} to overcome the context limitations of current methods. Instead of relying solely on the immediate dialogue, the proposed method leverages historical user experiences stored in an item recommendation graph, dynamically incorporating relevant contextual information into a base prompt that captures both past data and the current user query.  A unified conversational search and recommendation framework was developed in \cite{10.1145/3269206.3271776} that actively clarifies user needs through a System Ask User Respond (SAUR) paradigm. \cite{10.5555/3327546.3327641} introduces REDIAL, a publicly available, large-scale dataset containing over 10,000 real-world movie recommendation conversations, addressing the lack of comprehensive dialogue data in this domain. From a domain perspective, our work introduces a novel security domain. Moreover, instead of limiting recommendations to items as in previous study, we expanded the paradigm to a brand new AI assistant system built on top of skills and developed a solution to integrate domain knowledge, context and personalization to generate suggestions.
 
\section{Methodology}
Our approach combines several key components to create a comprehensive dynamic prompt recommendation system for domain-specific AI applications. Figure 1 illustrates the architecture of our system.

\subsection{System Architecture}
The system consists of several interdependent components, each designed to address a specific stage in the prompt recommendation pipeline. The core modules are:
\begin{itemize}
    \item \textbf{Contextual Query Processor:} Parses the user’s natural language query and enriches it with session, history, and user profile context.
    \item \textbf{Knowledge Retrieval Engine:} Leveraging a retrieval-augmented generation (RAG) approach to extract relevant knowledge, plugins, skills and supporting data from a comprehensive repository of domain-specific knowledge.
    \item \textbf{Hierarchical Skill Organization:} Structures skills into logical groups, enabling efficient selection of relevant plugins and skills.
    \item \textbf{Skill Ranking Engine:} Dynamically ranks candidate skills using predictive models trained on behavioral telemetry and user interaction history, ensuring that the most effective and contextually appropriate skills are prioritized for the current query.
    \item \textbf{Information Synthesis and Prompt Generation:} Combines all contextual and ranked information to generate meta-prompts, which are used to produce tailored prompt suggestions.
\end{itemize}
Each of these components is detailed in the following subsections, outlining their specific design and implementation within the system. 
\subsection{Contextual Query Processor}
When a user submits a natural language (NL) query to the AI application, the first component of our recommendation system to process this input is the Contextual Query Processor. This module ingests not only the raw NL query, but also the current session information, conversation history, and a user profile that include role, organizational context, and historical preferences. The processor parses the NL query to extract intent, relevant entities, and any domain-specific terminology. It encodes the session state by capturing recent dialogue turns, user feedback, and system actions. By integrating user profile data, the processor contextualizes the query, enabling downstream components to tailor their processing to the user's background and needs. The output of this stage is a structured, context-enriched representation of the user's query.
\subsection{Knowledge Retrieval Engine}
The next stage is the Knowledge Retrieval Engine, which is responsible for grounding the user’s query in relevant domain knowledge. The engine accesses a comprehensive knowledge repository that includes domain-specific documentation, data sources, and, most importantly, a hierarchical organization of available skills. To identify the most pertinent information, the engine employs a Retrieval-Augmented Generation (RAG) approach. Using semantic similarity between the query representation and knowledge base entries, the engine retrieves relevant information, plugins, individual skills, and supporting data sources. This retrieval process is context-aware, leveraging both the session context and broader organizational knowledge to ensure that only actionable and authorized information is surfaced for further processing.
\subsection{Hierarchical Skill Organization}
A distinguishing feature of our system is its hierarchical organization of skills. Rather than maintaining a flat list, skills are grouped into thematically related plugins, each containing multiple granular skills (More information regarding plugins and skills can be found here \footnote{\url{https://techcommunity.microsoft.com/blog/securitycopilotblog/how-to-become-a-microsoft-security-copilot-ninja-the-complete-level-400-training/4106928}}). This hierarchical structure enables a two-step reasoning process. During knowledge retrieval, the system first identifies the most relevant plugins in response to the user’s query and context. Subsequently, within each selected plugin, the system narrows down to the most relevant individual skills. This narrowing process is conceptually similar to the schema refinement step in NL2KQL~\cite{tang2024nl2kql}, which selects the most relevant schema elements to improve translation accuracy in domain-specific query systems. This approach not only improves computational efficiency but also enhances the precision of prompt recommendations by focusing on the most contextually appropriate capabilities.
\subsection{Skill Ranking Engine}
After relevant skills have been identified, the Skill Ranking Engine determines their prioritization for prompt generation. This module, informed by a predictive model trained on behavioral telemetry data, dynamically ranks skills based on both short-term session interactions and long-term usage patterns. Behavioral telemetry is collected through batch jobs that aggregate user interactions, skill invocation frequencies, and feedback across sessions. The predictive model leverages this data to infer which skills are most likely to be useful for the current user and context. The output is a ranked list of skills that balances immediate relevance with historical effectiveness for similar queries and users.
\subsection{Information Synthesis and Prompt Generation}
The final stage of the pipeline is the Information Synthesis and Prompt Generation module. This component aggregates all contextual signals—including the session context, conversation history, relevant knowledge snippets, and the ranked skill list—into a meta-prompt. The synthesis process employs both predefined and dynamically generated prompt templates, ensuring that the generated suggestions are clear, actionable, and aligned with organizational standards. To further enhance prompt quality, the system incorporates few-shot learning, injecting examples from similar historical queries to guide the generative AI model. The resulting meta-prompt is then passed to a large language model, which produces a set of candidate prompt suggestions tailored to the user’s intent and the domain’s operational constraints.

\section{Experiments}

We conducted comprehensive experiments to evaluate the effectiveness of our dynamic prompt suggestion system across different models and configurations.

\subsection{Experimental Setup}

\textbf{Dataset:} We analyzed data samples from opt-in customer sessions of a commercial, domain-specific AI assistant designed for security applications. Our evaluation dataset comprises 784 unique sessions totaling 2,967 chats, with each session featuring roughly three to four rounds of conversation on average. In these chats, a total of 12,432 prompts were suggested (with a hard cutoff of no more than five suggestions per chat). All suggestions underwent automated evaluation. Additionally, 152 chats were randomly and evenly selected from six major plugins for manual evaluation by security researchers.

\textbf{Models:} Three different model configurations were Compared:
\begin{itemize}
    \item \textbf{GPT-4o full pipeline:} GPT-4o for plugin inference, skill inference, and prompt generation
    \item \textbf{GPT-4o-mini hybrid:} GPT-4o-mini for plugin and skill inference with GPT-4o for prompt generation
    \item \textbf{Markov hybrid:} Markov model for plugin and skill inference with GPT-4o for prompt generation
\end{itemize}

\textbf{Evaluation Metrics:} A rubric-based evaluation framework were applied to assess the quality of the suggestions. Five key metrics including Relevance, Clarity, Novelty, Grounding and Usefulness were proposed with their corresponding detailed rubrics. The main concept of those metrics were list below. 
\begin{itemize}
    \item \textbf{Relevance:} A metric that evaluates how relevant a suggested prompt is to the conversation context
    \item \textbf{Clarity:} evaluating the clarity of a suggested prompt. The prompt must be clear and unambiguous, ensuring no room for misinterpretation
    \item \textbf{Novelty:} A measure of whether a suggestion is novel.The same prompt, or a slightly rephrased version, has not already been presented, and the answer is not available in the conversation history.
    \item \textbf{Grounding:} A merit of how well the suggested prompt aligns with conversation history. The suggestion does not contradict the conversation history, and all entities in prompt are either present in the historical conversation or relevant to it
    \item \textbf{Overall usefulness:} A measure of overall usefulness of the suggested prompt for completing user tasks described in the conversation history 
\end{itemize}

Due to page constraints, here’s an example rubric for overall usefulness.
\begin{itemize}
    \item \textbf{Level 0 - Completely Useless:} The prompt diverges from user tasks and provides no relevant information and does not connect with the conversation history, thereby failing to assist the user in any meaningful way.
    \item \textbf{Level 1 - Somewhat Useful:} The prompt provides some relevant information and offers partial assistance. While it might not be directly related to the conversation history, it still seeks information that can help the customer complete tasks or expand their activities. unambiguous, ensuring no room for misinterpretation.
    \item \textbf{Level 2 – Highly Useful:} The prompt suggests key information that is directly relevant to the user's tasks. It significantly helps the user in completing their tasks efficiently and effectively. 
\end{itemize}

\subsection{Automated Evaluation Results}

The automated evaluation of 12,432 suggested prompts showed strong performance across all model configurations, with an average usefulness score of 88.4\% for the full GPT-4o pipeline. Please note that the Grounding score appears slightly lower because some prompts use entity references, which results in lower scores. However, they are actually preferred as they make the prompts more concise and readable by replacing lengthy entity names with shorter references. This approach is intentional and aligns with our instructions for prompt generation.

\begin{table}[t]
\centering
\caption{Automated Evaluation Results}
\begin{adjustbox}{width=0.5\textwidth}
\begin{tabular}{lccccc}
\toprule
Model Configuration & Novelty & Grounding & Usefulness & Clarity & Relevance \\
\midrule
GPT-4o Full Pipeline & 0.933 & 0.772 & 0.884 & 0.865 & 0.863 \\
GPT-4o-mini Hybrid   & 0.905 & 0.770 & 0.870 & 0.829 & 0.845 \\
Markov + GPT-4o      & 0.920 & 0.804 & 0.885 & 0.897 & 0.851 \\
\bottomrule
\end{tabular}
\end{adjustbox}
\end{table}

The GPT-4o full pipeline demonstrated superior performance in novelty and relevance, while the Markov + GPT-4o configuration showed strengths in grounding and clarity. All configurations achieved similar overall usefulness scores, indicating the robustness of our approach across different model configurations.

\subsection{Manual Evaluation Results}

To validate the automated metrics, we conducted a manual evaluation with security researchers assessing 152 samples across different plugins including USX, Intune, MDTi, Purview, Entra, and NL2KQL. Unlike automated evaluation, manual evaluation assesses the entire set of suggestions (up to five) as a whole for each sample rather than evaluating each individual suggestion.

\begin{table}[t]
\centering
\caption{Manual Evaluation Results by Security Experts}
\begin{adjustbox}{width=0.5\textwidth}
\begin{tabular}{lccc}
\toprule
Model Configuration & Usefulness (\%) & Extremely Useful (\%) & Not Useful (\%) \\
\midrule
GPT-4o Full Pipeline & 98.0 & 75.0 & 2.0 \\
GPT-4o-mini Hybrid   & 96.5 & 53.0 & 3.5 \\
Markov + GPT-4o      & 98.9 & 53.1 & 1.1 \\
\bottomrule
\end{tabular}
\end{adjustbox}
\end{table}

Manual evaluation confirmed the effectiveness of our approach, with all configurations achieving usefulness rates above 96\%. The GPT-4o full pipeline demonstrated particular strength in generating ``extremely useful'' prompts (75\%), significantly outperforming other configurations in this category.

\subsection{Performance Analysis}

The experimental results demonstrate several key findings:
\begin{itemize}
    \item \textbf{High overall effectiveness:} All model configurations achieved strong usefulness scores in both automated and manual evaluations, confirming the robustness of our architectural approach.
    \item \textbf{Expert validation:} The high usefulness ratings from security researchers (96.5--98.9\%) validate that our system generates prompts that domain experts consider valuable.
    \item \textbf{Model trade-offs:} While the GPT-4o full pipeline excelled in generating extremely useful prompts, the Markov + GPT-4o configuration showed slightly higher overall usefulness with fewer ``not useful'' suggestions.
    \item \textbf{Domain adaptability:} Strong performance across multiple products (i.e., six plugins) demonstrates the system's effectiveness in diverse security contexts.
\end{itemize}

These results confirm that our dynamic prompt recommendation system successfully addresses the challenges of domain-specific prompt generation, offering significant improvements over existing approaches.
\section{Discussion}
We evaluated several models with an emphasis on cost efficiency. GPT-4o-mini, for example, is more than ten times cheaper than GPT-4o \footnote{\url{https://openai.com/api/pricing/}}. Our experiments revealed that both GPT-4o and GPT-4o-mini performed well in tasks related to identifying and inferring relevant plugins and skills. However, GPT-4o-mini performed much worse than GPT-4o on the prompt generation (around 15\% drop). We believe the discrepancy arises because the tasks of plugin and skill inference are relatively straightforward. In contrast, prompt generation requires not only determining which skills to employ, but also adhering to example prompts, diversifying the output, incorporating entity references and so on—all of which add complexity. Furthermore using Markov chain model, we leveraged behavioral telemetry data to infer relevant skills without the need for language model calls, thereby reducing costs significantly. However, this approach is effective only for popular, frequently used skills and does not resolve the cold start problem. To address this, we implemented a hybrid solution combining a language model with a Markov model, which effectively tackles cold starts while keeping costs low. One limitation to note is that many low-cost language models and potential alternatives to the Markov model have not yet been tested.

Diversity is a major challenge for recommendation systems, especially when it comes to open prompt suggestions. Our approach—which uses a skill-first strategy, step-by-step reasoning, and example prompts(Some example prompts could be found here \footnote{\url{https://learn.microsoft.com/en-us/copilot/security/prompting-security-copilot}} for your information)—is designed to generate varied recommendations. For instance, given the initial prompt "show me the recent signins" and its corresponding response, our system can offer a range of diverse suggestions: ["prompt": "List suspicious activities related to these sign-ins", "skill": "GetSecurityAlerts"],["prompt": "Show me the audit logs for these users","skill": "GetAuditLogs"], ["prompt": "Get all risky users involved in these sign-ins","skill": "GetRiskyUsers"],["prompt": "Show me the summary of risk for user Tom Monton", "skill": "GetRiskyUserSummary"],["prompt": "List risky Service Principals involved in these sign-ins", "skill": GetRiskyServicePrincipals"]. Unfortunately, we do not have a well-defined metric for that yet, which presents a potential direction for future research.

As mentioned earlier, prompt engineering is a significant challenge for AI systems built around a set of skills. Prompts that are off-topic (i.e., out-of-scope) or over-complex aren’t feasible. The key for a recommendation system is ensuring that the suggestions are feasible—meaning the skill-based AI system can correctly link a prompt to the appropriate skill and execute it effectively. This is quite challenge for open prompt suggestion systems. To address this challenge, our skill-first, example-based prompt generation strategy limits prompts to only those skills within scope and adheres to a clear example, thereby avoiding unnecessary complexity. Due to data access issues and privacy concerns, We haven’t developed specific metrics or conducted experiments to directly measure feasibility. An indirect test was conducted and compared the suggested skill against the skill predicted by the AI system, and we achieved an alignment of over 60\%. Although this figure might seem modest, it is quite reasonable given that skill prediction accuracy is 78\%.

\section{Conclusion and Future work}
In this paper, we presented a dynamic context-aware prompt recommendation system designed for domain-specific AI applications, with a primary focus on security operations. Our approach integrates contextual query processing, retrieval-augmented knowledge grounding, hierarchical skill organization, and adaptive skill ranking to generate prompt suggestions that are both actionable and tailored to user needs. The experiments demonstrated that our system achieves high performance across multiple evaluation metrics and model configurations. 

Despite these advances, several challenges remain. Our findings indicate that while lightweight models offer cost efficiency for skill inference, they are less effective at generating diverse and contextually rich prompts compared to larger language models. The hybrid approach, combining language models with statistical methods and behavioral telemetry, helps address the cold start problem and balances efficiency with recommendation quality. However, the measurement of prompt diversity and feasibility remains an open area for further investigation.

Looking ahead, future work will focus on several key directions. First, we aim to develop robust quantitative metrics for prompt diversity and feasibility, enabling more systematic evaluation and optimization of recommendation quality. Second, we plan to explore a broader range of low-cost language models and alternative ranking strategies to further improve efficiency and scalability. Third, addressing privacy and data access constraints will be essential for enabling richer personalization and behavioral modeling. Finally, we intend to generalize and validate our methodology across other specialized domains, such as healthcare and finance, to assess its adaptability and broader impact.

In summary, our research demonstrates that combining contextual awareness, domain knowledge, hierarchical reasoning, and adaptive personalization enables effective and scalable prompt recommendation in complex, skill-based AI systems. We believe these insights will inform the development of next-generation AI assistants and recommendation engines across a wide range of professional domains.

\begin{acks}
The authors gratefully recognize the following contributions to this work. Xinye led and coordinated the research and publication process, and was primarily responsible for the manuscript preparation and writing. Haijun proposed solutions, conducted experiments and evaluations, and assisted with manuscript writing. Chaitanya, Vineeth and Phillip developed the Markov model, and supported manuscript writing. Yogesh provided project guidance, strategic direction, and valuable insights throughout.

In addition to the contributions of the authors, we extend our sincere thanks to those whose support and expertise were instrumental to this work. In particular, we thank security expert Jessen Kurien for conducting all manual quality evaluations and providing extensive feedback. We also acknowledge Sylvie Liu for her support and for facilitating connections with our partners; Gurpreet Sohal and Ravi Teja Bellam for their insightful feedback and thoughtful concerns; and Kyle Johnson and Adam Hunt for their assistance in setting up the experimental environment. Finally, we express our gratitude to Yoni Nave, Steve Ginty, Ed Martin, Priyanka Tyagi, Luca Spolidoro, Hani Neuvirth-Telem, and Inbar Rotem for their valuable product feedback.
\end{acks}

\bibliographystyle{ACM-Reference-Format}
\bibliography{dps}


\begin{thebibliography}{19}

%%% the Web address.

\ifx \showCODEN    \undefined \def \showCODEN     #1{\unskip}     \fi
\ifx \showISBNx    \undefined \def \showISBNx     #1{\unskip}     \fi
\ifx \showISBNxiii \undefined \def \showISBNxiii  #1{\unskip}     \fi
\ifx \showISSN     \undefined \def \showISSN      #1{\unskip}     \fi
\ifx \showLCCN     \undefined \def \showLCCN      #1{\unskip}     \fi
\ifx \shownote     \undefined \def \shownote      #1{#1}          \fi
\ifx \showarticletitle \undefined \def \showarticletitle #1{#1}   \fi
\ifx \showURL      \undefined \def \showURL       {\relax}        \fi
% invisible to TeX
\providecommand\bibfield[2]{#2}
\providecommand\bibinfo[2]{#2}
\providecommand\natexlab[1]{#1}
\providecommand\showeprint[2][]{arXiv:#2}

\bibitem[Askari et~al\mbox{.}(2025)]%
        {askari2025magic}
\bibfield{author}{\bibinfo{person}{Arian Askari}, \bibinfo{person}{Christian Poelitz}, {and} \bibinfo{person}{Xinye Tang}.} \bibinfo{year}{2025}\natexlab{}.
\newblock \showarticletitle{Magic: Generating self-correction guideline for in-context text-to-sql}. In \bibinfo{booktitle}{\emph{Proceedings of the AAAI Conference on Artificial Intelligence}}, Vol.~\bibinfo{volume}{39}. \bibinfo{pages}{23433--23441}.
\newblock


\bibitem[Christakopoulou et~al\mbox{.}(2016)]%
        {10.1145/2939672.2939746}
\bibfield{author}{\bibinfo{person}{Konstantina Christakopoulou}, \bibinfo{person}{Filip Radlinski}, {and} \bibinfo{person}{Katja Hofmann}.} \bibinfo{year}{2016}\natexlab{}.
\newblock \showarticletitle{Towards Conversational Recommender Systems}. In \bibinfo{booktitle}{\emph{Proceedings of the 22nd ACM SIGKDD International Conference on Knowledge Discovery and Data Mining}} (San Francisco, California, USA) \emph{(\bibinfo{series}{KDD '16})}. \bibinfo{publisher}{Association for Computing Machinery}, \bibinfo{address}{New York, NY, USA}, \bibinfo{pages}{815–824}.
\newblock
\showISBNx{9781450342322}
\href{https://doi.org/10.1145/2939672.2939746}{doi:\nolinkurl{10.1145/2939672.2939746}}


\bibitem[Lester et~al\mbox{.}(2021)]%
        {DBLP:journals/corr/abs-2104-08691}
\bibfield{author}{\bibinfo{person}{Brian Lester}, \bibinfo{person}{Rami Al{-}Rfou}, {and} \bibinfo{person}{Noah Constant}.} \bibinfo{year}{2021}\natexlab{}.
\newblock \showarticletitle{The Power of Scale for Parameter-Efficient Prompt Tuning}.
\newblock \bibinfo{journal}{\emph{CoRR}}  \bibinfo{volume}{abs/2104.08691} (\bibinfo{year}{2021}).
\newblock
\showeprint[arXiv]{2104.08691}
\urldef\tempurl%
\url{https://arxiv.org/abs/2104.08691}
\showURL{%
\tempurl}


\bibitem[Lewis et~al\mbox{.}(2020)]%
        {lewis2020retrieval}
\bibfield{author}{\bibinfo{person}{Patrick Lewis}, \bibinfo{person}{Ethan Perez}, \bibinfo{person}{Aleksandra Piktus}, \bibinfo{person}{Fabio Petroni}, \bibinfo{person}{Vladimir Karpukhin}, \bibinfo{person}{Naman Goyal}, \bibinfo{person}{Heinrich K{\"u}ttler}, \bibinfo{person}{Mike Lewis}, \bibinfo{person}{Wen-tau Yih}, \bibinfo{person}{Tim Rockt{\"a}schel}, {et~al\mbox{.}}} \bibinfo{year}{2020}\natexlab{}.
\newblock \showarticletitle{Retrieval-augmented generation for knowledge-intensive nlp tasks}.
\newblock \bibinfo{journal}{\emph{Advances in neural information processing systems}}  \bibinfo{volume}{33} (\bibinfo{year}{2020}), \bibinfo{pages}{9459--9474}.
\newblock


\bibitem[Li et~al\mbox{.}(2018)]%
        {10.5555/3327546.3327641}
\bibfield{author}{\bibinfo{person}{Raymond Li}, \bibinfo{person}{Samira Kahou}, \bibinfo{person}{Hannes Schulz}, \bibinfo{person}{Vincent Michalski}, \bibinfo{person}{Laurent Charlin}, {and} \bibinfo{person}{Chris Pal}.} \bibinfo{year}{2018}\natexlab{}.
\newblock \showarticletitle{Towards deep conversational recommendations}. In \bibinfo{booktitle}{\emph{Proceedings of the 32nd International Conference on Neural Information Processing Systems}} (Montr\'{e}al, Canada) \emph{(\bibinfo{series}{NIPS'18})}. \bibinfo{publisher}{Curran Associates Inc.}, \bibinfo{address}{Red Hook, NY, USA}, \bibinfo{pages}{9748–9758}.
\newblock


\bibitem[Liu et~al\mbox{.}(2021)]%
        {DBLP:journals/corr/abs-2107-13586}
\bibfield{author}{\bibinfo{person}{Pengfei Liu}, \bibinfo{person}{Weizhe Yuan}, \bibinfo{person}{Jinlan Fu}, \bibinfo{person}{Zhengbao Jiang}, \bibinfo{person}{Hiroaki Hayashi}, {and} \bibinfo{person}{Graham Neubig}.} \bibinfo{year}{2021}\natexlab{}.
\newblock \showarticletitle{Pre-train, Prompt, and Predict: {A} Systematic Survey of Prompting Methods in Natural Language Processing}.
\newblock \bibinfo{journal}{\emph{CoRR}}  \bibinfo{volume}{abs/2107.13586} (\bibinfo{year}{2021}).
\newblock
\showeprint[arXiv]{2107.13586}
\urldef\tempurl%
\url{https://arxiv.org/abs/2107.13586}
\showURL{%
\tempurl}


\bibitem[Ma et~al\mbox{.}(2023a)]%
        {ma-etal-2023-dynamic}
\bibfield{author}{\bibinfo{person}{Xuan Ma}, \bibinfo{person}{Tieyun Qian}, {and} \bibinfo{person}{Ke Sun}.} \bibinfo{year}{2023}\natexlab{a}.
\newblock \showarticletitle{Dynamic Open-book Prompt for Conversational Recommender System}. In \bibinfo{booktitle}{\emph{Findings of the Association for Computational Linguistics: EMNLP 2023}}, \bibfield{editor}{\bibinfo{person}{Houda Bouamor}, \bibinfo{person}{Juan Pino}, {and} \bibinfo{person}{Kalika Bali}} (Eds.). \bibinfo{publisher}{Association for Computational Linguistics}, \bibinfo{address}{Singapore}, \bibinfo{pages}{9839--9849}.
\newblock
\href{https://doi.org/10.18653/v1/2023.findings-emnlp.658}{doi:\nolinkurl{10.18653/v1/2023.findings-emnlp.658}}


\bibitem[Ma et~al\mbox{.}(2023b)]%
        {ma2023dynamic}
\bibfield{author}{\bibinfo{person}{Xuan Ma}, \bibinfo{person}{Tieyun Qian}, {and} \bibinfo{person}{Ke Sun}.} \bibinfo{year}{2023}\natexlab{b}.
\newblock \showarticletitle{Dynamic Open-book Prompt for Conversational Recommender System}. In \bibinfo{booktitle}{\emph{Findings of the Association for Computational Linguistics: EMNLP 2023}}. \bibinfo{pages}{9839--9849}.
\newblock


\bibitem[Mao et~al\mbox{.}(2025)]%
        {mao2025reinforcedpromptpersonalizationrecommendation}
\bibfield{author}{\bibinfo{person}{Wenyu Mao}, \bibinfo{person}{Jiancan Wu}, \bibinfo{person}{Weijian Chen}, \bibinfo{person}{Chongming Gao}, \bibinfo{person}{Xiang Wang}, {and} \bibinfo{person}{Xiangnan He}.} \bibinfo{year}{2025}\natexlab{}.
\newblock \bibinfo{title}{Reinforced Prompt Personalization for Recommendation with Large Language Models}.
\newblock
\showeprint[arxiv]{2407.17115}~[cs.IR]
\urldef\tempurl%
\url{https://arxiv.org/abs/2407.17115}
\showURL{%
\tempurl}


\bibitem[Ranaldi et~al\mbox{.}(2025)]%
        {ranaldi2025multilingual}
\bibfield{author}{\bibinfo{person}{Leonardo Ranaldi}, \bibinfo{person}{Barry Haddow}, {and} \bibinfo{person}{Alexandra Birch}.} \bibinfo{year}{2025}\natexlab{}.
\newblock \showarticletitle{Multilingual Retrieval-Augmented Generation for Knowledge-Intensive Task}.
\newblock \bibinfo{journal}{\emph{arXiv preprint arXiv:2504.03616}} (\bibinfo{year}{2025}).
\newblock


\bibitem[Sander~Schulhoff(2025)]%
        {schulhoff2025promptreportsystematicsurvey}
\bibfield{author}{\bibinfo{person}{Michael Ilie et~al. Sander~Schulhoff}.} \bibinfo{year}{2025}\natexlab{}.
\newblock \bibinfo{title}{The Prompt Report: A Systematic Survey of Prompt Engineering Techniques}.
\newblock
\showeprint[arxiv]{2406.06608}~[cs.CL]
\urldef\tempurl%
\url{https://arxiv.org/abs/2406.06608}
\showURL{%
\tempurl}


\bibitem[Schick and Sch{\"{u}}tze(2020)]%
        {DBLP:journals/corr/abs-2001-07676}
\bibfield{author}{\bibinfo{person}{Timo Schick} {and} \bibinfo{person}{Hinrich Sch{\"{u}}tze}.} \bibinfo{year}{2020}\natexlab{}.
\newblock \showarticletitle{Exploiting Cloze Questions for Few-Shot Text Classification and Natural Language Inference}.
\newblock \bibinfo{journal}{\emph{CoRR}}  \bibinfo{volume}{abs/2001.07676} (\bibinfo{year}{2020}).
\newblock
\showeprint[arXiv]{2001.07676}
\urldef\tempurl%
\url{https://arxiv.org/abs/2001.07676}
\showURL{%
\tempurl}


\bibitem[Shin et~al\mbox{.}(2020)]%
        {shin-etal-2020-autoprompt}
\bibfield{author}{\bibinfo{person}{Taylor Shin}, \bibinfo{person}{Yasaman Razeghi}, \bibinfo{person}{Robert~L. Logan~IV}, \bibinfo{person}{Eric Wallace}, {and} \bibinfo{person}{Sameer Singh}.} \bibinfo{year}{2020}\natexlab{}.
\newblock \showarticletitle{{A}uto{P}rompt: {E}liciting {K}nowledge from {L}anguage {M}odels with {A}utomatically {G}enerated {P}rompts}. In \bibinfo{booktitle}{\emph{Proceedings of the 2020 Conference on Empirical Methods in Natural Language Processing (EMNLP)}}, \bibfield{editor}{\bibinfo{person}{Bonnie Webber}, \bibinfo{person}{Trevor Cohn}, \bibinfo{person}{Yulan He}, {and} \bibinfo{person}{Yang Liu}} (Eds.). \bibinfo{publisher}{Association for Computational Linguistics}, \bibinfo{address}{Online}, \bibinfo{pages}{4222--4235}.
\newblock
\href{https://doi.org/10.18653/v1/2020.emnlp-main.346}{doi:\nolinkurl{10.18653/v1/2020.emnlp-main.346}}


\bibitem[Tang et~al\mbox{.}(2024)]%
        {tang2024nl2kql}
\bibfield{author}{\bibinfo{person}{Xinye Tang}, \bibinfo{person}{Amir~H Abdi}, \bibinfo{person}{Jeremias Eichelbaum}, \bibinfo{person}{Mahan Das}, \bibinfo{person}{Alex Klein}, \bibinfo{person}{Nihal~Irmak Pakis}, \bibinfo{person}{William Blum}, \bibinfo{person}{Daniel~L Mace}, \bibinfo{person}{Tanvi Raja}, \bibinfo{person}{Namrata Padmanabhan}, {et~al\mbox{.}}} \bibinfo{year}{2024}\natexlab{}.
\newblock \showarticletitle{NL2KQL: From Natural Language to Kusto Query}.
\newblock \bibinfo{journal}{\emph{arXiv preprint arXiv:2404.02933}} (\bibinfo{year}{2024}).
\newblock


\bibitem[Wei et~al\mbox{.}(2022)]%
        {Weidblp2022}
\bibfield{author}{\bibinfo{person}{Jason Wei}, \bibinfo{person}{Xuezhi Wang}, \bibinfo{person}{Dale Schuurmans}, \bibinfo{person}{Maarten Bosma}, \bibinfo{person}{Ed~H. Chi}, \bibinfo{person}{Quoc Le}, {and} \bibinfo{person}{Denny Zhou}.} \bibinfo{year}{2022}\natexlab{}.
\newblock \showarticletitle{Chain of Thought Prompting Elicits Reasoning in Large Language Models}.
\newblock \bibinfo{journal}{\emph{CoRR}}  \bibinfo{volume}{abs/2201.11903} (\bibinfo{year}{2022}).
\newblock
\showeprint[arXiv]{2201.11903}
\urldef\tempurl%
\url{https://arxiv.org/abs/2201.11903}
\showURL{%
\tempurl}


\bibitem[Xu et~al\mbox{.}(2020)]%
        {xu-etal-2020-user}
\bibfield{author}{\bibinfo{person}{Hu Xu}, \bibinfo{person}{Seungwhan Moon}, \bibinfo{person}{Honglei Liu}, \bibinfo{person}{Bing Liu}, \bibinfo{person}{Pararth Shah}, \bibinfo{person}{Bing Liu}, {and} \bibinfo{person}{Philip Yu}.} \bibinfo{year}{2020}\natexlab{}.
\newblock \showarticletitle{User Memory Reasoning for Conversational Recommendation}. In \bibinfo{booktitle}{\emph{Proceedings of the 28th International Conference on Computational Linguistics}}, \bibfield{editor}{\bibinfo{person}{Donia Scott}, \bibinfo{person}{Nuria Bel}, {and} \bibinfo{person}{Chengqing Zong}} (Eds.). \bibinfo{publisher}{International Committee on Computational Linguistics}, \bibinfo{address}{Barcelona, Spain (Online)}, \bibinfo{pages}{5288--5308}.
\newblock
\href{https://doi.org/10.18653/v1/2020.coling-main.463}{doi:\nolinkurl{10.18653/v1/2020.coling-main.463}}


\bibitem[Zhang et~al\mbox{.}(2018)]%
        {10.1145/3269206.3271776}
\bibfield{author}{\bibinfo{person}{Yongfeng Zhang}, \bibinfo{person}{Xu Chen}, \bibinfo{person}{Qingyao Ai}, \bibinfo{person}{Liu Yang}, {and} \bibinfo{person}{W.~Bruce Croft}.} \bibinfo{year}{2018}\natexlab{}.
\newblock \showarticletitle{Towards Conversational Search and Recommendation: System Ask, User Respond}. In \bibinfo{booktitle}{\emph{Proceedings of the 27th ACM International Conference on Information and Knowledge Management}} (Torino, Italy) \emph{(\bibinfo{series}{CIKM '18})}. \bibinfo{publisher}{Association for Computing Machinery}, \bibinfo{address}{New York, NY, USA}, \bibinfo{pages}{177–186}.
\newblock
\showISBNx{9781450360142}
\href{https://doi.org/10.1145/3269206.3271776}{doi:\nolinkurl{10.1145/3269206.3271776}}


\bibitem[Zhao et~al\mbox{.}(2021)]%
        {DBLP:journals/corr/abs-2102-09690}
\bibfield{author}{\bibinfo{person}{Tony~Z. Zhao}, \bibinfo{person}{Eric Wallace}, \bibinfo{person}{Shi Feng}, \bibinfo{person}{Dan Klein}, {and} \bibinfo{person}{Sameer Singh}.} \bibinfo{year}{2021}\natexlab{}.
\newblock \showarticletitle{Calibrate Before Use: Improving Few-Shot Performance of Language Models}.
\newblock \bibinfo{journal}{\emph{CoRR}}  \bibinfo{volume}{abs/2102.09690} (\bibinfo{year}{2021}).
\newblock
\showeprint[arXiv]{2102.09690}
\urldef\tempurl%
\url{https://arxiv.org/abs/2102.09690}
\showURL{%
\tempurl}


\bibitem[Zhou et~al\mbox{.}(2023)]%
        {zhou2023efficientpromptingdynamicincontext}
\bibfield{author}{\bibinfo{person}{Wangchunshu Zhou}, \bibinfo{person}{Yuchen~Eleanor Jiang}, \bibinfo{person}{Ryan Cotterell}, {and} \bibinfo{person}{Mrinmaya Sachan}.} \bibinfo{year}{2023}\natexlab{}.
\newblock \bibinfo{title}{Efficient Prompting via Dynamic In-Context Learning}.
\newblock
\showeprint[arxiv]{2305.11170}~[cs.CL]
\urldef\tempurl%
\url{https://arxiv.org/abs/2305.11170}
\showURL{%
\tempurl}


\end{thebibliography}

\end{document}